\documentclass{article}
\usepackage[utf8]{inputenc} \usepackage{graphicx, times}

\title{Anomaly detection in laser-guided vehicles' batteries: a case study}
\author{Gianfranco Lombardo (1), Stefano Cagnoni (1), Stefano Cavalli (1), \\ Juan Jos\'e Contreras Gonz\'ales (2), Francesco Monica (2), \\ Monica Mordonini (1), Michele Tomaiuolo (1)\\ \\ (1) Dept. of Engineering and Architecture, University of Parma \\ (2) Elettric80 spa, Reggio Emilia}
\date{}

\begin{document}
\maketitle
\begin{abstract}
Detecting anomalous data within time series is a very relevant task in pattern recognition and machine learning, with many possible applications that range from disease prevention in medicine, e.g., detecting early alterations of the health status before it can clearly be defined as "illness" up to monitoring industrial plants. Regarding this latter application, detecting anomalies in an industrial plant's status firstly prevents serious damages that would require a long interruption of the production process. Secondly, it permits optimal scheduling of maintenance interventions by limiting them to urgent situations. At the same time, they typically follow a fixed prudential schedule according to which components are substituted well before the end of their expected lifetime.
This paper describes a case study regarding the monitoring of the status of Laser-guided Vehicles (LGVs) batteries, on which we worked as our contribution to project SUPER (Supercomputing Unified Platform, Emilia Romagna) aimed at establishing and demonstrating a regional High-Performance Computing platform that is going to represent the main Italian supercomputing environment for both computing power and data volume.

\end{abstract}

\section{Introduction}
The term {\em anomaly detection}, with regard to time series analysis, has been long used in statistics, even if not so frequently, as a possible alternative for the more common term {\em outlier detection} \cite{kozma1994anomaly}.\\
Recently, more and more frequently, such a term has been referred to the analysis of time series coming from sensors describing the health status of industrial plants. Within this context, it has rapidly gained relevance as a research and real-life topic. This is partially due the actions, taken within by the Industry 4.0 framework, that have encouraged industrial plant connectivity and the use of sensor networks and Internet of Things (IoT) devices for real-time monitoring of properties describing a plant's status. \\
In a productive process, the first concern is indeed to guarantee that plants are always in a correct working status, since the cost of systematic maintenance, even when performed according to a non-optimal, over-frequent schedule, is lower than the cost of suspending production following an unexpected (and possibly catastrophic) fault. 

Therefore, accurately monitoring their plants' status allows companies to define models that may then be applied to plants' maintenance from two different viewpoints:  

\begin{itemize}
    \item {\em anomaly detection}, i.e., a timely detection of a possible sudden malfunctioning before it produces an outburst of problems that may damage the plant and cause an expensive long-term suspension of the production process;

    \item {\em remaining lifetime estimation}, i.e., an estimate of the remaining time before the probability of a fault occurring becomes higher than a preset threshold.
\end{itemize} 

Monitoring data from the above two viewpoints is the basis for the optimization of a plant's maintenance schedule, usually termed {\em predictive maintenance} (PM). 
In the absence of long-term dense monitoring data, PM can only be based on empirical knowledge that derives from a macroscopic and generally incomplete plant description. Indeed, defining a robust model of the plant's behavior requires a long time; in the meanwhile, maintenance has to be cautiously performed at more or less prefixed intervals that are much shorter than the expected lifetime of a plant's critical components.

Even if the term {\em predictive maintenance} has been used for more than twenty years \cite{levitt2003complete,mobley2002introduction}, the models to which such a term refers nowadays are much different than the ones that were used originally. This equally depends on the different quality and amount of data that can be acquired from plants and on the recent developments of machine-learning models in terms of quality, intrinsic power, and capability of dealing with huge amounts of complex data. In fact, the vast majority of PM applications nowadays rely on machine-learning methods \cite{carvalho2019machine,ccinar2020machine, mo2021evolutionary}, even if one should not be surprised noticing that, in spite of very different limitations, \cite{kozma1994anomaly} already dealt with neural network-based models.

However, in most cases, the huge amount of data currently available is very far from ideal from a machine-learning (ML) perspective, especially when dealing with binary classification problems like the one that we will consider in the case study described in the following sections.\\
From this viewpoint, theoretical and practical rules suggest that, in solving a binary classification
problem, an optimal training data set for an ML application should include
roughly as many positive as negative samples, i.e., it should be what is usually
termed a {\em balanced} data set. To obtain data satisfying such a requirement, it would be obviously too costly and impractical to replicate complex large industrial plants in a laboratory to study their behavior under extreme stress conditions up to the occurrence of faults that often imply the destruction of some of their components. Because of this, the only data available for building machine learning models can be those acquired from the plants during the actual production cycle. During the latter, before any {\em a posteriori} optimization of the maintenance schedule can be applied, the plant is kept under a prudential (and costly) maintenance schedule, which relies on a strong underestimation of the remaining lifetime of components to avoid virtually any anomalous event.
Thus, even in the presence of an infrastructure that would allow data about faults to be acquired, the data acquired from the sensors almost exclusively relate to healthy plants, which makes it almost impossible to characterize possible
anomalies in the plant based on classical supervised-learning approaches.\\
The most common way for ML methods to deal with this problem is therefore by building a model of the healthy plant (usually referred to as a {\em digital twin} of the plant) and compare the real data coming from the plant's sensor network to the corresponding data produced by the digital twin over a certain time interval. The interpretation of the results of such a comparison assumes that a small divergence between the two time series corresponds to a healthy status of the plant, while the occurrence of larger discrepancies may suggest that something is probably going wrong with it.
 
Within a larger regional project on Big Data analysis (SUPER \-- "Supercomputing Unified Platform, Emilia Romagna"), we investigated the effectiveness of these machine learning-based analysis methodologies by studying, applying, and validating some state-of-the-art Deep Learning techniques \cite{goodfellow2016deep} applied to the automatic analysis of large amounts of data from an industrial plant.

\section{A real-world test case}
Aiming at the aforementioned goals, we analyzed the data provided by Elettric80 spa, partner of the project. The dataset they provided is the result of a data ingestion process conducted by the company in order to record and monitor various functional parameters of their LGVs (Laser Guided Vehicles) that operate in a water-bottling plant.

The available data relate to 25 LGVs and were collected on a daily basis in the period between January 2020 and March 2021 for about ten hours each day. In particular, for each robot, the data refer to:

\begin{itemize}
\item Vehicle system variables (status, current operation)
\item Battery: temperatures, state of charge, charge cycles, amperage and voltage
\item Vehicle navigation: LGV position relative to environmental markers
\item Mast of the automatic forklift: weight on board, height reached, estimate of the chain elongation
\item Tilt: (lifting cylinder pressure, working time, etc.)
\item Hydraulic pumps for lifting: (inverter temperature, speed setpoint, torque feedback, etc.)
\item Steering: (inverter temperature, motor speed, etc.)

\end{itemize}

The dataset allows one to deal with different forecasting tasks; we focused on identifying anomalies in the temperature of lithium batteries since this study was of particular interest to the company. In fact, these batteries may malfunction due to time wear but also due to their overheating if the air filters of the LGV are clogged by dust typically affecting industrial plants. As a result, a model has been developed that can automatically detect anomalous temperature behaviors and act as alarms urging for maintenance and extraordinary cleaning.

\section{Automatic detection of anomalies in the temperature of lithium batteries}

The dataset does not contain additional "external" information that labels as normal or anomalous the data acquired within the temporal windows into which the dataset was divided. In fact, we consider the data sequence acquired within a time window of $250~s$ as a sample to be classified. 
According to this organization, we designed an architecture for data analysis within which a neural network-based autoencoder was trained to generate a model of the normal data. Such a model is used as a reference to identify anomalous data (i.e., those data that cannot be correctly replicated by the model) in a totally unsupervised way. 
The autoencoder receives a sequence/sample as input and then compresses it by encoding the useful information as the output of a hidden layer much smaller than the input layer, performin what can be termed {\em unsupervised feature learning}. The self-supervised autoencoder architecture is completed by adding a further hidden layer, symmetric to the one used in the encoder part of the network, and an output layer having the same size as the input layer. The full network is then trained to reproduce the input signal (i.e., the data acquired from the plant) in the output layer (signal reconstruction).
If this type of architecture is trained by examples of data acquired in normal working conditions for the plant, it will likely commit a small error in replicating these signals while failing to reconstruct, or at least approximate as precisely, anomalous signals. This behavior can be explained considering that the network has learned the probability distribution of 'normal' data, and therefore has no knowledge of 'anomalous' samples having a different distribution, to which it reacts in some unpredictable way.

Two different neural-network architectures were tested to implement the autoencoder: a Multi-Layer Perceptron (MLP) and a GRU (Gated-Recurrent Unit) recurrent neural network. This design choice is justified by the results achieved in the state of the art by these neural models  \cite{Liufaultrecurrent, Loyseqgru,Provotarlstmautoenc}. In particular, the choice of recurrent networks like the GRU and LSTM (Long Short Term Memory) networks aims to exploit their memory and ability to embed time dependencies.  In fact, we also tested LSTMs but they confirmed some documented adverse computational effects, like gradient explosion/vanishing \cite{hochreitervanishing98,explovanishribeiro20}. With the data we considered, problems concerning the overflow of the values of the network weights did occur. Although they could be solved using gradient-clipping techniques, we could not achieve the same satisfactory results we obtained using the GRUs, at the cost of the latter’s higher computational requirements. 
From a computational efficiency viewpoint, we also considered a more traditional MLP-based approach \cite{haykinneuralnets09} for its lower computational burden when processing big data, in view of its subsequent use on a much greater amount of data than we used in our experimentation.\\

{\noindent \em Operational pipeline \\}

{\noindent The} investigation activity with both architectures was conducted according to the following pipeline:

\begin{enumerate}
\item {\em Data cleaning}. We identified and corrected data inconsistencies and deal with missing data.
\item {\em Identification of the operating scenario}. A common operating scenario was identified for all LGVs in order to exclude some elements that may alter the temperature: transporting particularly heavy or bulky packages, long journeys. The best scenario was identified in the battery recharging phases, since all LGVs are frequently in this phase, in similar conditions and positions within the working environment. In fact, all the batteries are recharged when they reach a percentage of remaining charge between $70$ and $80\%$. Each robot is then recharged several times a day.
\item {\em Data analysis}. The batteries are recharged until $90\%$ of the full charge is reached; this requires a time, depending on the starting conditions, in the order of minutes.
\item {\em Time series processing}. Since the data is sampled every $250~ms$, each recharge phase has been divided into time windows $1000$ samples long with an overlap of $50\%$. The size of this window was identified through a parameter optimization process based on a grid search.
\item {\em Dataset creation}. The available data has been divided into a training and a test set on a time basis. The training set comprises data from year $2020$ while the test set those acquired during $2021$.
\item {\em Parameter optimization and choice}. We have taken a validation set apart from the training based on which we chose the parameters that define an anomaly (i.e., the reconstruction error threshold).
\item {\em Analysis of the anomalies detected}.

\end{enumerate}

\section{Data Processing with GRU networks}

GRU-type recurring networks have the main advantage of exhibiting memory properties; they can therefore operate directly on time series by processing one sample at a time based on the current input value but also on the values of the time sequence previously processed. In our use-case, the input time series is represented by the sampling of the temperature of the LGV batteries being recharged within a pre-defined time window. Each instance is made up of $1000$ consecutive samples of the variable acquired within a $250 s$ long time window.

The autoencoder we designed to create the data model uses two separate GRU networks: one for the encoding (learning of a representation for the time series) and one for the decoding (reconstruction of the input signal as network output) operations. In the design phase, recurrent networks of different lengths in terms of memory cells were evaluated to study the model dependence on this parameter, i.e., on a lower or higher compression of the information in the encoding phase. Specifically, networks with $128, 256, 512,$ and $1024$ GRU cells were tested.

\begin{figure}[h!]

\centering
{\includegraphics[width=0.80\textwidth]{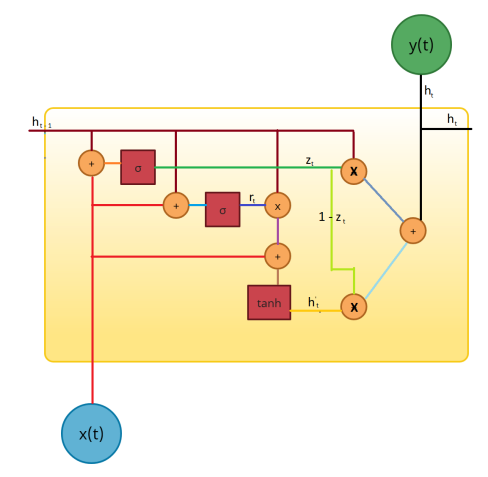}}%
\hfill
\caption{GRU- Gated Recurrent Unit}
\label{fig:Figure1}
\end{figure}

Each GRU unit corresponds to a single time step, so it receives the corresponding sample in input along with the output from the previous unit or the initialization vector (h) to limit the network depth (that could theoretically be unlimited) and thus regulates the "forgetting" dynamics of the effects of the partial time sequence processed up to that point.
One unit of this model implements: an addition operation (+), a sigmoid function ($\sigma$), a Hadamard product operation (x) and a {\em tanh} function. In practice, GRUs are capable of storing and filtering information using their unique update and reset gates. This eliminates the vanishing gradient problem, i.e., the fading of the gradient value propagated back during training with the increase of the network depth, since the model retains, within its state, memory of the previously analyzed data, finally transmitting them to the unit corresponding to the next time step.

In order to identify the best architecture capable of minimizing the signal reconstruction error using GRU networks, we proceeded with the training of different models in which the main training parameters were the same, except for the number of memory cells:

\begin{itemize}
\item Learning rate = $0.001$
\item Batch size = 32
\item Number of epochs = 32
\end{itemize}

The networks were compared on a validation set using the Loss function used for training, i.e., the Mean-Squared Error (MSE) between the input time sequence and the reconstructed (output) sequence (Figure 2).

\begin{figure}[h!]

\centering
{\includegraphics[width=\textwidth]{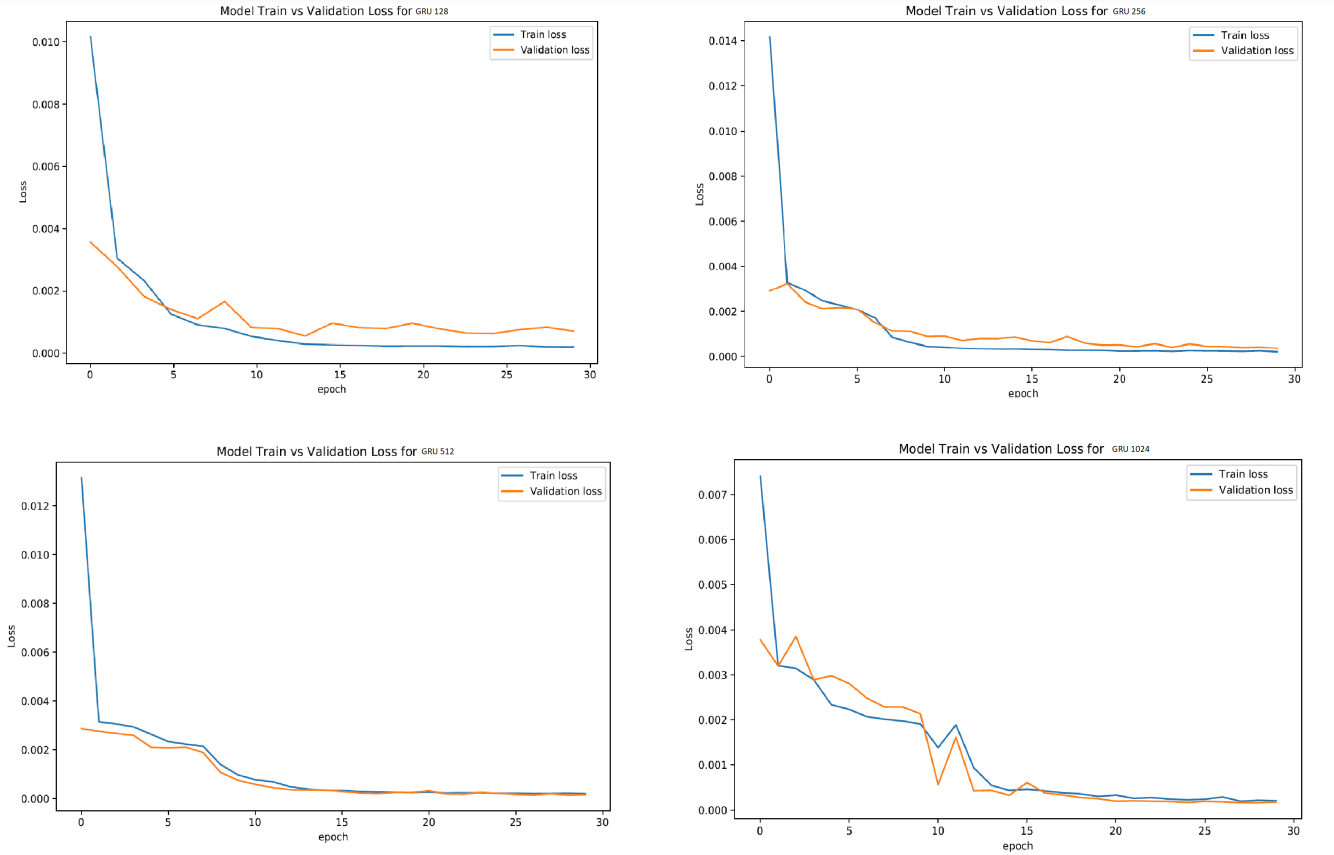}}%
\hfill
\caption{Comparison of the results on the training set and on the validation set.}
\label{fig:Figure2}
\end{figure}

Subsequently, this result was used to analyze the distribution of the error between the samples of the validation and the test set in order to determine the error threshold beyond which a temporal sequence should be identified as anomalous based on the reconstruction error (Figure 3).

\begin{figure}[h!]

\centering
{\includegraphics[width=0.90\textwidth]{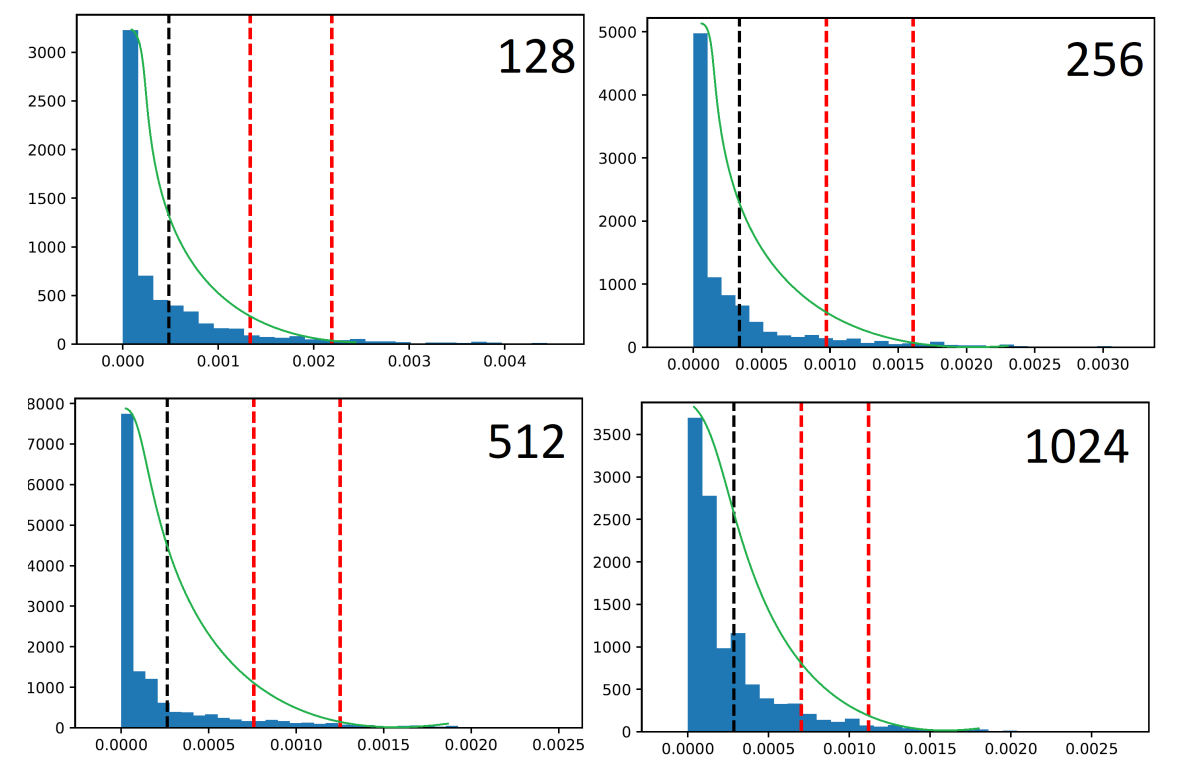}}%
\hfill
\caption{Analysis of the distribution of the reconstruction error (MSE) with different numbers of GRU cells. The average error on the training set is marked in black, while the possible thresholds, identified as multiples of the standard deviation of data are marked in red.}
\label{fig:Figure3}
\end{figure}

The best results were obtained with the model with $1024$ GRU units. Setting an MSE threshold equal to $0.0007$, it identifies as anomalous about $9.66\%$ of the samples in the validation set and about $10.19\%$ of the test set.

\section{Processing data with a Multi-Layer Perceptron}

Recurrent networks have proven to be excellent tools for identifying anomalies, while still showing some computational efficiency problems with large datasets. For this reason, we have also tested an architecture based on Multi-Layer Perceptrons in order to address larger training sets with a limited computational burden.

Differently from recurrent networks, MLPs do not process time series with architectures forming a chain of units, each corresponding to a different acquisition time. Instead, the entire input sequence is fed into the input layer as a single instance (a one-dimensional vector), ignoring the reproduction of the sequence dynamics. The network design was therefore constrained by the size of the time sequences ($1000$ samples). The structure of the autoencoder therefore comprises an input layer of $1000$ neurons and a hidden layer of $200$ neurons in the encoding section; a layer of $40$ neurons as the internal representation (embedding) of the input data; finally, a hidden layer of $200$ neurons and an output layer of $1000$ neurons in the decoding section.

The reconstruction error was measured as the mean square error between the vector of the input samples and the output vector reconstructed in the output, averaged over all the time sequences in the validation set  and, subsequently, in the test set to determine the optimal error threshold for identifying anomalies.

The reconstruction error is distributed according to a Gaussian with mean value $0.176$ and variance $0.02$. The threshold that determines the maximum reconstruction error before considering the sample an anomaly, was set to $0.432$ as the average value of the reconstruction error in validation plus three times the standard deviation. This choice is based on the confidence intervals of the Gaussian; in particular, with this choice $99\%$ of the distribution centered on the average value is preserved and only the most 'critical' instances are identified as anomalies.

\begin{figure}[h!]

\centering
{\includegraphics[width=\textwidth]{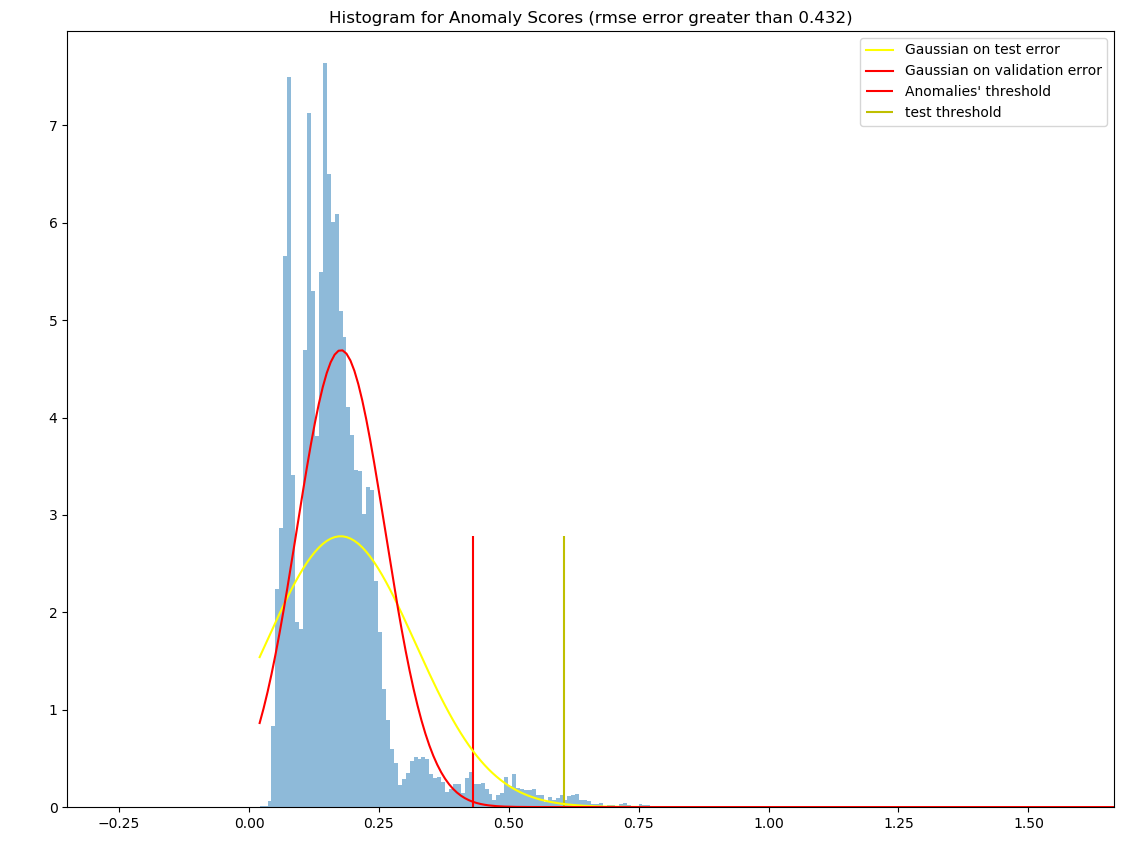}}%
\hfill
\caption{Reconstruction error on the test set overlapped to the Gaussian distribution best fit on the  validation set (in red) and the corresponding threshold (red bar)}
\label{fig:Figure4}
\end{figure}

Using this threshold, we could identify $548$ anomalies in the test set comprising all data acquired in 2021. We analyzed these data using K-means clustering \cite{macqueen1967kmeans} and adding Dynamic Time Warping (DTW) \cite{muller2007dynamic} for a more robust calculation of the Euclidean distance between the samples and the corresponding centroids. The number of clusters is equal to 3. Finally, these clusters were analyzed using Principal Component Analysis (PCA) \cite{abdi2010principal} and T Stochastic Neighborhood Embedding (T-SNE) \cite{van2008tsne}. Figure 5 displays the clustering results.

\begin{figure}[h!]

\centering
{\includegraphics[width=\textwidth]{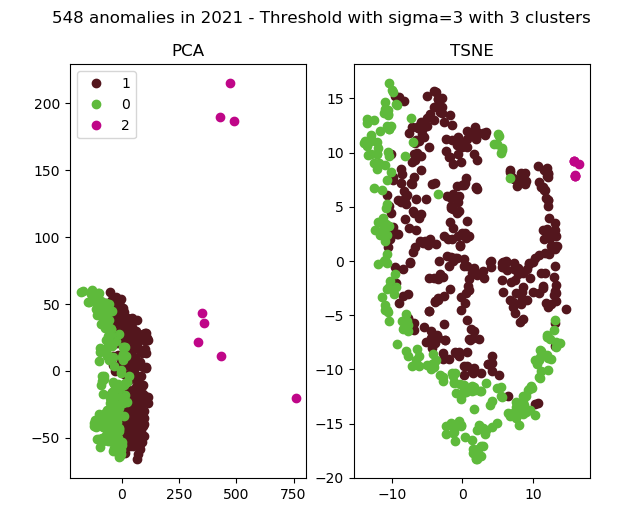}}%
\hfill
\caption{Clustering of anomalies.}
\label{fig:Figure5}
\end{figure}

We analyzed the three clusters that can be identified by observing the corresponding anomalies in the time domain appreciating the good separation obtained by our unsupervised approach and hypothesizing the possible causes for such a cluster distribution. This analysis made it possible to formulate a hypothesis for each cluster:

\begin{itemize}
\item Cluster 0: These are all the time windows within which one can observe a sudden drop in temperature,  whose average is about 5 degrees centigrade. Corresponding to these events, the battery stops absorbing current for a few seconds (the value of current going from negative to zero in figure 3). The battery power suddenly drops by an average of 10 percent and then it starts recharging again.
\item Cluster 1: The samples appear to be very similar to those of cluster 0. However, the "anomalous" time windows seem to correspond to the end of the LGV batteries recharge.
\item Cluster 2: All these time windows show an abnormal and sudden rise in temperature at the end of which the battery begins to absorb current and the remaining charge begins to increase. The anomalous condition seems to affect the LGV batteries in the initial moments of recharging, as shown in Figure 4.
\end{itemize}

\begin{figure}[h!]
\begin{tabular}{c c}
{\includegraphics[width=0.47\textwidth]{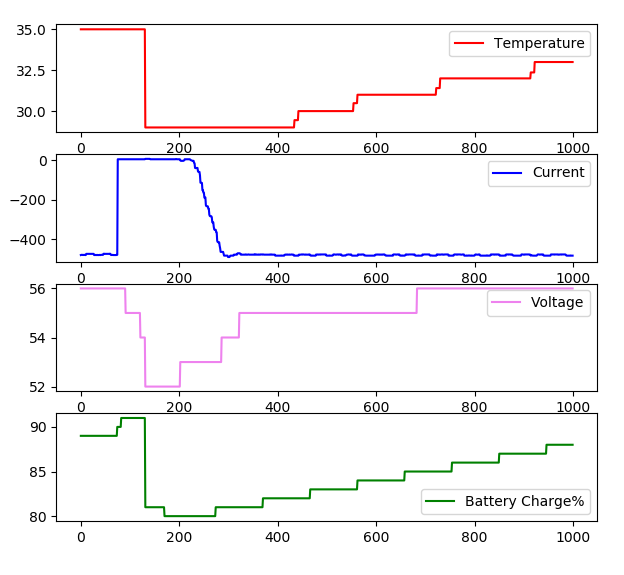}} &
{\includegraphics[width=0.405\textwidth]{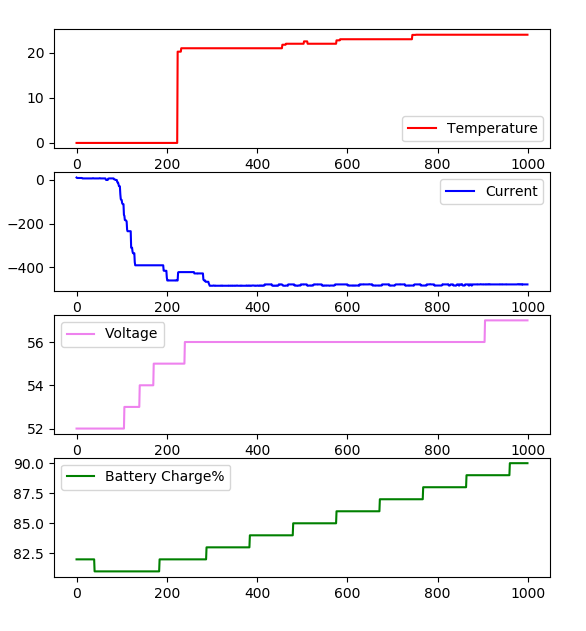}}\\
\end{tabular}
\hfill

\caption{A sample from cluster 0 (left) and a sample from cluster 2 (right)}
\label{fig:Figure67}
\end{figure}

In conclusion, using the MLP-based model with the identified threshold, it was possible to discriminate some anomalies in the behavior of the battery. However, further investigations need to be carried out directly on the LGV in its working environment to correctly identify the causes that lead to the occurrence of such situations.

\section{Conclusions and future developments}

The investigation carried out so far has led to promising results which, however, require further investigation. The unavailability of supervised data only permits a limited validation of the model. However, we could observe that the anomalous time windows reported by the system actually correspond to clearly identifiable situations. The anomaly detection system we designed therefore accomplishes its task.

A future step could involve the collection of data from an LGV, in its operating environment, in which the clogging of the filters would be artificially caused. The consequent temperature data acquisition could offer the chance to obtain further validation data by which we might determine a causal relationship for the anomalies. In particular, considering a scenario in which the LGVs are in regular operation, in addition to the charging phase, the same 'anomalous' behaviors could be observed in the robots' braking phase. This situation therefore appears to be worth further investigation in the future, as braking is an opportunity for recovering energy and recharging the batteries on board.

The future availability of data in much greater quantities, when the acquisition system we considered is extended to all the plants where the LGVs are in operation and, in any case, new data can also be acquired from the plants already under monitoring, will also allow to proceed with further analyses, the most important of which could be the estimate of the remaining life time (RUL) of lithium batteries. This parameter is crucial in view of current and future applications of these batteries in other different domains. 

\section*{Acknowledgments}
Gianfranco Lombardo and Stefano Cavalli were supported by a grant from Regione Emilia Romagna (SUPER \-- "Supercomputing Unified Platform, Emilia Romagna", Azione 1.5.1, POR-FESR 2014-2020 Emilia Romagna).

\bibliography{pmaint.bib}
\bibliographystyle{splncs03.bst}

\end{document}